\documentclass[pdflatex,sn-mathphys-num]{sn-jnl}

\usepackage{graphicx}
\usepackage{amsmath,amssymb,amsfonts}
\usepackage[table]{xcolor}
\usepackage{booktabs}
\usepackage{listings}

\usepackage{geometry}
\makeatletter\@twosidefalse\@mparswitchfalse\makeatother

\usepackage{pifont}

\definecolor{checkgreen}{HTML}{2E7D5B}
\definecolor{crossred}{HTML}{9B2C2C}
\newcommand{\yes}{\textcolor{checkgreen}{\ding{51}}}
\newcommand{\no}{\textcolor{crossred}{\ding{55}}}
\newcommand{\partyes}{\textcolor{checkgreen!70!black}{\textasciitilde}}

\lstdefinestyle{pythonstyle}{
  language=Python,
  basicstyle=\ttfamily\footnotesize,
  keywordstyle=\color{blue!70!black}\bfseries,
  stringstyle=\color{green!45!black},
  commentstyle=\color{gray}\itshape,
  backgroundcolor=\color{gray!10},
  frame=single,
  framerule=0pt,
  rulecolor=\color{gray!10},
  showstringspaces=false,
  breaklines=true,
  columns=fullflexible,
  aboveskip=6pt,
  belowskip=6pt,
  xleftmargin=1em,
  framexleftmargin=1em,
  framexrightmargin=0.5em,
  framextopmargin=4pt,
  framexbottommargin=4pt,
}
\lstdefinelanguage{json}{
  basicstyle=\ttfamily\footnotesize,
  commentstyle=\color{gray}\itshape,
  stringstyle=\color{green!45!black},
  morestring=[b]",
  morecomment=[l]{//},
  literate=
    {:}{{{\color{blue!70!black}{:}}}}{1}
    {,}{{{\color{blue!70!black}{,}}}}{1}
    {\{}{{{\color{blue!70!black}{\{}}}}{1}
    {\}}{{{\color{blue!70!black}{\}}}}}{1}
    {[}{{{\color{blue!70!black}{[}}}}{1}
    {]}{{{\color{blue!70!black}{]}}}}{1}
    {null}{{{\color{red!60!black}{null}}}}{4}
    {true}{{{\color{red!60!black}{true}}}}{4}
    {false}{{{\color{red!60!black}{false}}}}{5},
}
\lstdefinestyle{jsonstyle}{
  language=json,
  backgroundcolor=\color{gray!10},
  frame=single,
  framerule=0pt,
  rulecolor=\color{gray!10},
  showstringspaces=false,
  breaklines=true,
  columns=fullflexible,
  aboveskip=6pt,
  belowskip=6pt,
  xleftmargin=1em,
  framexleftmargin=1em,
  framexrightmargin=0.5em,
  framextopmargin=4pt,
  framexbottommargin=4pt,
}
\begin{document}

\title[RadHarmony]{RadHarmony: Radiological Data Handling in the Era of Agentic AI}

\author[1]{\fnm{Frank} \sur{Li}}
\author[2]{\fnm{Bardia} \sur{Khosravi}}
\author[1]{\fnm{Mohammadreza} \sur{Chavoshi}}
\author[1]{\fnm{Theo} \sur{Dapamede}}
\author[1]{\fnm{YoungSeok} \sur{Jeon}}
\author[1]{\fnm{Janice} \sur{Newsome}}
\author[1]{\fnm{Hari} \sur{Trivedi}}
\author*[1]{\fnm{Judy} \sur{Gichoya}}\email{judywawira@emory.edu}

\affil*[1]{\orgdiv{Department of Radiology and Imaging Sciences}, \orgname{Emory University}, \orgaddress{\city{Atlanta}, \state{GA}, \country{USA}}}

\affil[2]{\orgdiv{Department of Radiology}, \orgname{Yale University}, \orgaddress{\city{New Haven}, \state{CT}, \country{USA}}}

\abstract{Training deep learning models on radiological images requires integrating heterogeneous datasets across different sources, file formats, directory layouts, label schemas, and annotation types. We present RadHarmony, an open-source Python library that provides a unified API for loading, harmonizing, and augmenting radiological datasets, with a primary focus on chest radiographs and early support for computed tomography (CT) and magnetic resonance imaging (MRI). RadHarmony standardizes metadata from 24 public datasets into a single tabular format, wraps MONAI's map-style datasets for deep-learning-ready sample delivery with optional on-disk caching, and supports classification labels, segmentation masks, bounding boxes, and radiology report text through a single interface, with an interactive visualization tool for dataset exploration and verification. To lower the barrier for integrating new datasets, RadHarmony introduces an AI-agent skill that guides the full integration workflow from raw data inspection through code generation and testing. We demonstrate the library's utility by pretraining RadHarmony-ViT, a reference vision transformer baseline that combines three heterogeneous chest radiograph datasets with no dataset-specific code. The code and pretrained model weights are available at \url{https://github.com/f10409/RadHarmony}.}


\maketitle

\section{Introduction}\label{sec:intro}

Large-scale radiological datasets such as CheXpert~\cite{ref1}, MIMIC-CXR~\cite{ref2}, ChestX-ray14~\cite{ref3}, and CT-RATE~\cite{ref4} have been instrumental in advancing deep learning for medical image analysis. However, each dataset is distributed with its own metadata CSV schema, directory structure, label vocabulary, and image format (DICOM, JPEG, PNG, NIfTI, or NumPy). As a consequence, researchers typically write custom data-loading code for each dataset. This process is error-prone, difficult to reproduce, and presents a significant barrier to multi-dataset training, cross-dataset evaluation, and data version control.

\begin{figure}[!htbp]
  \centering
  \includegraphics[width=\linewidth]{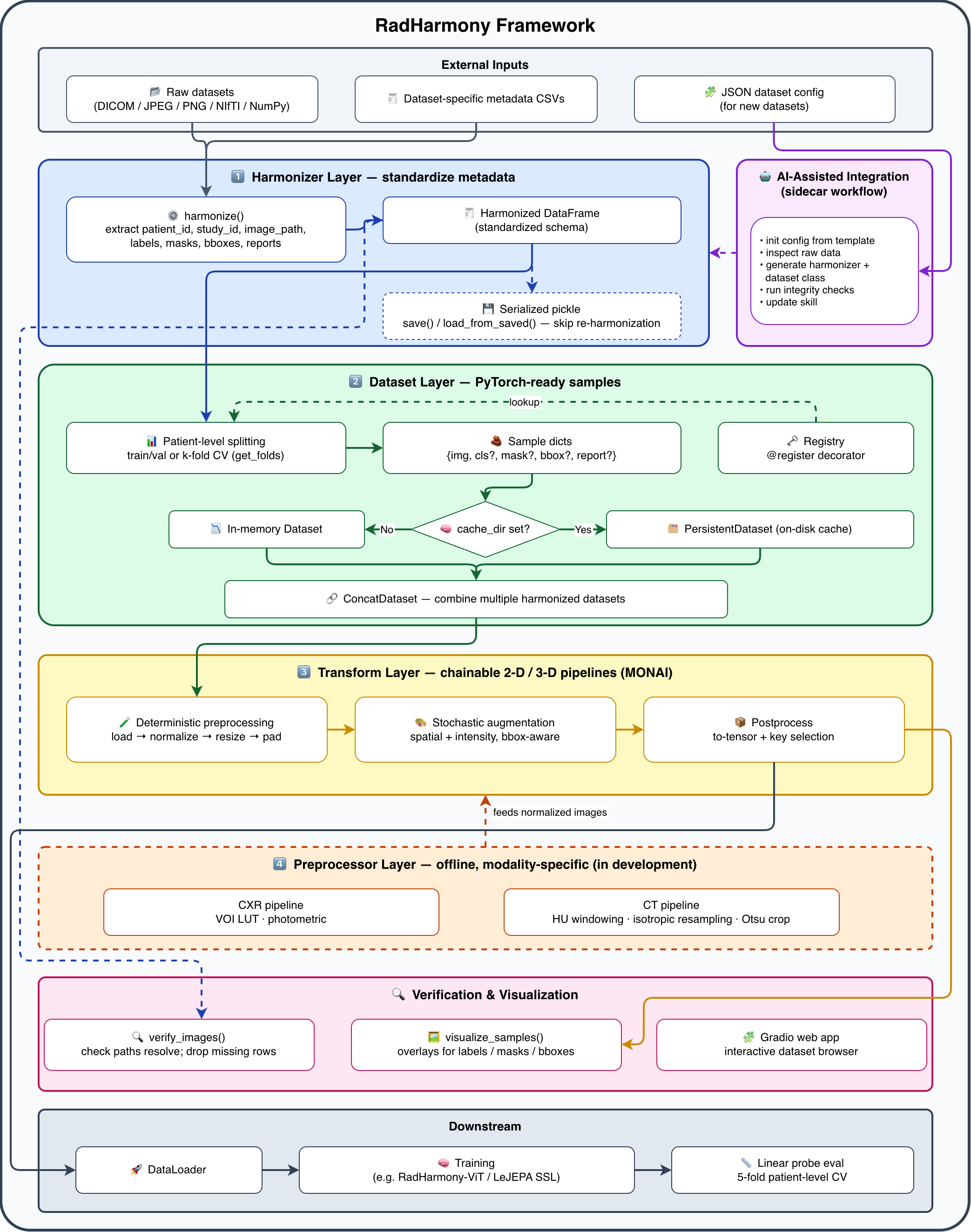}
  \caption{Overview of the RadHarmony pipeline. The harmonizer layer standardizes dataset-specific metadata into a common schema; the dataset layer builds MONAI-cached PyTorch datasets with patient-level splits; and the transform layer composes deterministic preprocessing with stochastic augmentation for 2-D and 3-D modalities.}\label{fig:pipeline}
\end{figure}

RadHarmony addresses the data harmonization challenge with a layered architecture comprising three principal components (Figure~\ref{fig:pipeline}): a \emph{harmonizer layer} that ingests raw, dataset-specific metadata and produces a standardized tabular representation; a \emph{dataset layer} that delivers PyTorch-ready sample dictionaries with built-in patient-level splitting and cross-validation; and a \emph{transform layer} that provides a chainable builder API for composing preprocessing and augmentation pipelines for both 2-D and 3-D modalities. RadHarmony does not redistribute or host any dataset. Instead, it provides an API that encodes the domain knowledge required to curate and wrangle publicly available radiological datasets into AI-ready form. Users download the original datasets themselves under each dataset's own license and terms; RadHarmony then harmonizes the metadata and loads images through a unified interface. Beyond the built-in datasets, users can integrate their own datasets through a guided tutorial notebook or the AI-assisted workflow.

The broader takeaways of this work are threefold. First, data harmonization is a prerequisite for scalable medical imaging research yet it remains largely unaddressed by existing tools; RadHarmony shows that encoding dataset-specific domain knowledge into a reusable harmonization layer eliminates this bottleneck and makes multi-dataset training as simple as single-dataset training. Second, AI-agent-assisted code generation, under human verification, is well-suited to convention-heavy software engineering in healthcare ML: a structured prompt paired with a declarative schema enables an AI coding agent to perform end-to-end dataset integration, offering a reusable recipe for keeping healthcare ML libraries extensible as new datasets and modalities emerge. Third, lowering the engineering barrier to multi-dataset training directly enables studies that were previously engineering-heavy to set up: our case study combines three chest radiograph datasets (spanning DICOM, JPEG, and PNG formats) for self-supervised pretraining with no dataset-specific code and evaluates the effect of additional data using supervised linear probes on a fourth external dataset.

\section{Related work}\label{sec:related}

Several efforts have sought to standardize medical image data handling. MONAI~\cite{ref5} provides composable, dictionary-based transforms and persistent caching for PyTorch pipelines, while TorchIO~\cite{ref6} offers similar functionality with a focus on 3-D medical volumes. Libraries such as MedMNIST~\cite{ref7} package curated subsets into standardized benchmarks, but are constrained to low-resolution images that are only marginally suitable for classification and inadequate for tasks such as segmentation or detection. TorchXRayVision~\cite{ref16} is the most closely related prior work: it provides a common interface and preprocessing chain for multiple chest X-ray datasets, enabling single-line dataset swapping and merged training. However, TorchXRayVision focuses on chest radiographs with bundled pretrained models. RadHarmony likewise centers on chest imaging in the current release but is designed as a modality-agnostic framework, with early support for 3-D CT and MRI datasets and a broader set of annotation types (segmentation masks, bounding boxes, and radiology reports) exposed through a single interface. Moreover, most of these tools operate at the transform or benchmark level and do not address the upstream challenge of harmonizing heterogeneous metadata (the mapping from dataset-specific metadata CSVs, label encodings, and path conventions into a single, consistent representation suitable for both training and evaluation). Table~\ref{tab:related} compares RadHarmony, MedMNIST and TorchXRayVision across different axes important for multi-dataset radiological experiments.

\begin{table}[!htbp]
  \centering
  \caption{Capability comparison of multi-dataset medical-imaging libraries. \yes{} = supported; \no{} = not supported; \partyes{} = partial or indirect support. MONAI and TorchIO are omitted as they are general-purpose transform/I-O toolkits rather than dataset-integration libraries; RadHarmony in fact builds on MONAI's caching and transform infrastructure.}\label{tab:related}
  \footnotesize
  \begin{tabular}{@{}lccc@{}}
    \toprule
    \textbf{Capability} & \textbf{MedMNIST}\cite{ref7} & \textbf{TorchXRayVision}\cite{ref16} & \textbf{RadHarmony} \\
    \midrule
    Metadata harmonization          & \no & \yes & \yes \\
    2-D imaging                     & \yes & \yes & \yes \\
    3-D imaging                     & \yes & \no & \yes \\
    Native-resolution images        & \no & \yes & \yes \\
    Classification labels           & \yes & \yes & \yes \\
    Segmentation masks              & \no & \yes & \yes \\
    Bounding boxes                  & \no & \partyes\footnotemark[1] & \yes \\
    Radiology reports               & \no & \no & \yes \\
    Patient-level splits / CV       & \partyes & \partyes & \yes \\
    Persistent on-disk caching      & \no & \no & \yes \\
    Interactive visualizer          & \no & \no & \yes \\
    AI-assisted dataset integration & \no & \no & \yes \\
    \botrule
  \end{tabular}
  \footnotetext[1]{TorchXRayVision exposes bounding-box annotations as binary pathology masks rather than as box coordinates.}
\end{table}

\section{Methods}\label{sec:methods}

\subsection{Harmonizer layer}\label{sec:harmonizer}

The harmonizer layer converts dataset-specific metadata CSV into a standardized tabular representation. Each dataset requires a harmonizer that defines how to extract three essential fields (patient identifier, study identifier, and relative image path) from the raw CSV columns. Optional hooks extract radiographic view position, classification labels, segmentation masks, bounding boxes, and radiology report text. When annotations reside in a separate file, the harmonizer automatically joins them on configurable key columns. The harmonizer also materializes annotations that are not stored as ready-to-load files. For example, SIIM-ACR ships pneumothorax masks as run-length-encoded strings; the harmonizer decodes them into PNG files and records the paths so downstream code always sees a uniform \texttt{mask\_path} column. A verification function checks that all image paths resolve to files on disk, dropping unresolvable rows to prevent downstream failures.

The harmonized output conforms to a fixed schema (Table~\ref{tab:schema}). Harmonized results can be serialized to disk, avoiding the need to repeat expensive operations such as DICOM header parsing or directory traversal. Code~\ref{lst:harmonizer} shows a minimal end-to-end invocation: instantiate the dataset-specific harmonizer with the raw metadata CSV, call \texttt{harmonize()}, and receive a standardized DataFrame ready for downstream dataset construction.

\begin{lstlisting}[caption={Running a harmonizer. The raw dataset-specific CSV is converted into the unified tabular schema of Table~\ref{tab:schema} with a single call.},label={lst:harmonizer},float=t]
from radharmony.harmonizer import CheXpertHarmonizer

harmonizer = CheXpertHarmonizer(csv_path="CheXpert-v1.0/train.csv")
df = harmonizer.harmonize(base_image_dir=CHEXPERT_DIR)
# df columns: patient_id, study_id, image_path, view_position,
#             atelectasis, cardiomegaly, ..., pleural_effusion
\end{lstlisting}

\begin{table}[!htbp]
  \centering
  \caption{Harmonized schema produced by the harmonizer layer.}\label{tab:schema}
  \begin{tabular}{@{}lll@{}}
    \toprule
    \textbf{Column} & \textbf{Type} & \textbf{Description} \\
    \midrule
    \texttt{patient\_id}    & str            & Unique patient identifier \\
    \texttt{study\_id}      & str            & Unique study or examination identifier \\
    \texttt{image\_path}    & str            & Relative path to the image file \\
    \texttt{view\_position} & str (optional) & Radiographic projection (AP, PA, lateral) \\
    \texttt{mask\_path}     & str (optional) & Path to segmentation mask \\
    \texttt{bbox}           & list (optional)& Normalized bounding box coordinates ($[0,1]$) \\
    \texttt{bbox\_labels}   & list (optional)& Labels associated with each bounding box \\
    \texttt{report}         & str (optional) & Radiology report text \\
    \texttt{<label\_col>}   & int (optional) & Binary label (0/1) per pathological finding \\
    \botrule
  \end{tabular}
\end{table}

\subsection{Dataset layer}\label{sec:dataset}

The dataset layer consumes harmonized metadata and delivers PyTorch-ready sample dictionaries built on MONAI's map-style datasets infrastructure. Each sample contains a normalized image tensor and optionally a classification label vector, segmentation mask, bounding box coordinates, and report text.

Two data-splitting strategies are provided: a single train/validation partition and a $k$-fold mode for cross-validation. Both operate at the patient level to prevent data leakage. The dataset constructor accepts raw metadata CSV paths, a pre-harmonized DataFrame, or a serialized harmonizer, decoupling the harmonization step from dataset creation for rapid iteration. Boolean output flags selectively enable annotation types, so a single dataset class can serve classification, segmentation, detection, and report generation tasks. Code~\ref{lst:kfold} illustrates how patient-level $k$-fold cross-validation is exposed as a single generator call.

\begin{lstlisting}[caption={Patient-level $k$-fold cross-validation. The dataset class yields \texttt{(train, val)} tuples per fold, with patients partitioned once and reused across folds to prevent leakage.},label={lst:kfold},float=t]
from radharmony.dataset import VinDrCXRTrainDataset
from torch.utils.data import DataLoader

ds = VinDrCXRTrainDataset(base_image_dir=VINDR_DIR, transform=tfm)
for fold, (train_ds, val_ds) in enumerate(ds.get_folds(n_splits=5)):
    train_loader = DataLoader(train_ds, batch_size=32, shuffle=True)
    val_loader   = DataLoader(val_ds,   batch_size=32)
    ...  # fit / evaluate on this fold
\end{lstlisting}

\subsection{Transform layer}\label{sec:transform}

The transform layer provides a chainable builder API for composing MONAI transform pipelines, with separate builders for 2-D and 3-D modalities. Pipelines are organized into three phases: deterministic preprocessing (loading, normalization, resizing, padding), stochastic augmentation (spatial and intensity transforms via chainable builder methods), and postprocessing (tensor conversion and key selection). Users may also supply their own MONAI transform pipeline, making the system flexible for custom preprocessing and augmentation requirements. Code~\ref{lst:transform} shows the chainable builder idiom used to compose a 2-D preprocessing and augmentation pipeline in a single expression.

\begin{lstlisting}[caption={Chainable transform builder for 2-D radiographs. Deterministic preprocessing (load, resize, pad) is configured via the constructor; stochastic augmentations are layered on through chained \texttt{with\_*} calls, and \texttt{add\_transform} provides an escape hatch for arbitrary MONAI transforms.},label={lst:transform},float=t]
import monai as mn
from radharmony.dataset import RadiologyTransform2D

tfm = (
    RadiologyTransform2D(img_size=256, output_keys={"img", "cls"})
    .with_flip(spatial_axis=1)
    .with_affine(translate_range=(10, 10), rotate_range=(0.17,))
    .with_intensity_jitter(shift=0.05)
    .add_transform(  # escape hatch for arbitrary MONAI transforms
        mn.transforms.RandGaussianNoiseD(keys=["img"], prob=0.2, std=0.01)
    )
    .get_transform()
)
\end{lstlisting}

Bounding box annotations are handled during spatial augmentation by converting normalized coordinates into a temporary binary mask, applying the same spatial transforms as the image, and recovering the transformed coordinates. This reuses the existing image transform pipeline to keep bounding boxes geometrically consistent, without requiring separate coordinate-tracking logic.

\subsection{Preprocessor layer}\label{sec:preprocessor}

An offline preprocessing layer (currently under active development) handles format-specific image preparation: DICOM VOI correction~\cite{Dapamede2025}, photometric interpretation normalization, foreground cropping via Otsu thresholding, and isotropic resampling. Separate preprocessing pipelines are provided for chest radiographs and CT volumes, each tailored to the modality's requirements.

\subsection{Registry and visualizer}\label{sec:registry}

A decorator-based registry enables datasets to be discovered and instantiated by string key at runtime. Third-party or custom datasets can register themselves by applying the decorator at import time, extending the library without modifying its source code. Code~\ref{lst:registry} illustrates both registration and lookup.

\begin{lstlisting}[caption={Dataset registry. Built-in datasets are registered via the \texttt{@register\_dataset} decorator; users can register custom datasets in the same way. At runtime, \texttt{resolve\_dataset} returns the class by string key.},label={lst:registry},float=t]
from radharmony.registry import register_dataset, resolve_dataset, list_datasets
from radharmony.dataset import BaseRadiologicalDataset

# Register a custom dataset
@register_dataset("my_dataset")
class MyDataset(BaseRadiologicalDataset):
    _HARMONIZER_CLS = MyDatasetHarmonizer
    ...

# Discover and instantiate by name
print(list_datasets())  # ['chexpert', 'mimic_cxr', 'my_dataset', ...]
DatasetCls = resolve_dataset("my_dataset")
ds = DatasetCls(base_image_dir="/data/my_dataset", output_cls=True)
\end{lstlisting}

\begin{figure}[!htbp]
  \centering
  \includegraphics[width=0.75\linewidth]{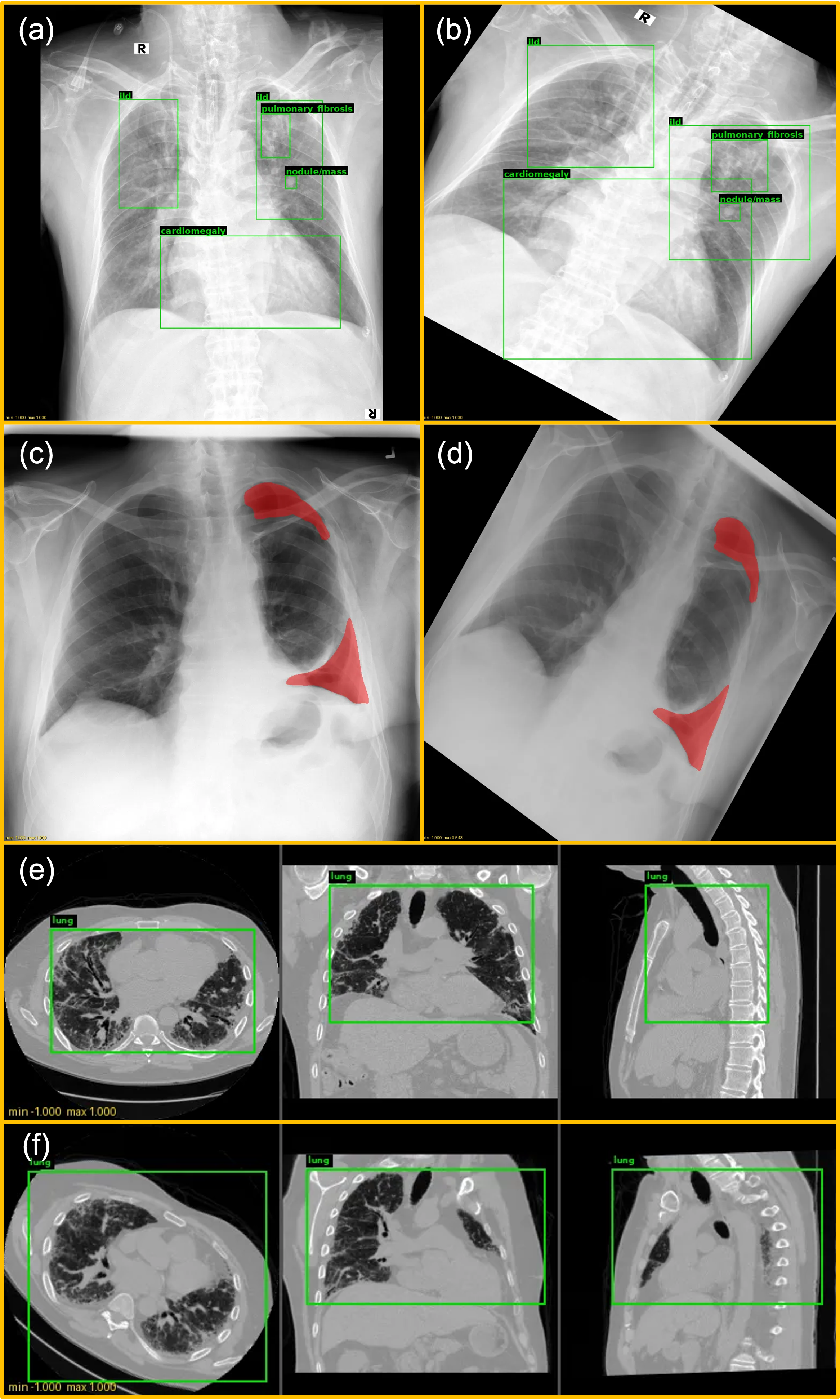}
  \caption{Interactive visualizer output. Original and augmented samples for VinDr-CXR (a,b; left/right), SIIM-ACR Pneumothorax (c,d; left/right), and RAD-ChestCT (e,f; top/bottom). Bounding boxes and segmentation masks are geometrically transformed together with the image, remaining consistent with the augmented pixels.}\label{fig:gradio}
\end{figure}

A Gradio-based~\cite{ref14} web application provides interactive dataset exploration and verification. The app consumes the same PyTorch datasets used during training, so the samples it renders are exactly what the model will see. Users select a dataset, configure annotation display options, and browse randomly sampled examples with live augmentation previews (Figure~\ref{fig:gradio}).

\subsection{AI-assisted dataset integration}\label{sec:ai-integration}

Integrating a new dataset requires writing multiple coordinated artifacts (harmonizer, dataset class, visualization wiring, and integrity checks) all conforming to strict framework conventions. This task is highly structured yet detail-sensitive, making it well-suited for AI agent automation.

The integration process is anchored by a JSON configuration file that declaratively describes how to read and interpret a dataset (where the files are, how to extract patient and study identifiers from metadata columns, which columns contain labels, and what annotation types are available). Code~\ref{lst:rsna-config} in the appendix shows an example configuration for the RSNA Pneumonia dataset. A companion Jupyter notebook guides users through inspecting their data and exporting a completed configuration.

Given this configuration, the AI agent executes a five-stage workflow: (1) initialize the configuration from a template, pre-filling fields inferred from the user's description; (2) inspect the raw data to fill remaining fields; (3) generate all code artifacts following the library's templates; (4) run automated verification, combining a static checklist for cross-file contracts (label-column ordering and harmonizer/dataset consistency, \texttt{\_HARMONIZER\_CLS} linkage, complete \texttt{\_\_init\_\_.py} exports, visualizer-registry entries) with a runtime integration test that adds the new dataset to a shared integrity notebook and runs \texttt{verify\_images()} (confirming every image path resolves to a readable file on disk) and \texttt{check\_dataset()} (iterating every sample through a PyTorch DataLoader and reporting decode failures, label-vector shape mismatches, and first-sample metadata such as image dtype, tensor shape, and label distribution), iteratively patching the offending code until both the checklist and the notebook cell pass; and (5) summarize the experience and update the skill definition if improvements are identified. The agent is instructed never to modify base classes, halting and notifying the user if such a change appears necessary. Of the 24 datasets currently supported by RadHarmony, three (MIMIC-CXR, CT-RATE, and SIIM-ACR PTX) were drafted by hand and later refined by the agent, serving as the reference templates from which the remaining 21 were integrated through human-agent collaboration using this workflow. As a practical test of the workflow's usability, the RSNA Pneumonia dataset~\cite{ref17} was integrated entirely by a physician user with no prior familiarity with the framework.

\subsection{Contribution workflow and continuous integration}\label{sec:contrib}

RadHarmony is developed as an open-source, community-driven project. Contributions (new datasets, transforms, and bug fixes) are submitted as pull requests on GitHub. Every push and pull request triggers a GitHub Actions workflow that runs a pytest suite covering harmonizer schemas, dataset instantiation, registry lookup, transform composition, and bounding-box handling. This CI acts as a repository-wide quality gate, enforcing the framework's contracts on every contribution.

\subsection{Case study: RadHarmony-ViT}\label{sec:case-study}

To demonstrate that RadHarmony's unified API facilitates multi-dataset experiments with minimal engineering overhead, we pretrain RadHarmony-ViT, a reference vision transformer baseline. RadHarmony-ViT is not intended as a final model but as a starting point that users can adapt to their own downstream tasks.

RadHarmony-ViT is pretrained with LeJEPA~\cite{ref15}, a self-supervised objective that aligns the embeddings of all augmented views toward the mean global-view embedding while regularizing the embedding distribution toward an isotropic Gaussian prior. The model uses a NaFlex ViT-Base backbone~\cite{ref18} (patch size 16, input resolution 256) with a multi-crop strategy: two global views and eight local views per image. Training uses AdamW with OneCycleLR (cosine annealing, 2.5\% warmup, peak LR $1\mathrm{e}{-4}$) for 100{,}000 steps at a batch size of 256 on 2 NVIDIA RTX Pro 6000 Blackwell GPUs with gradient checkpointing.

Training data is drawn from three chest radiograph datasets simultaneously (CheXpert~\cite{ref1}, MIMIC-CXR~\cite{ref2}, and ChestX-ray14~\cite{ref3}), loaded and harmonized through RadHarmony. Images carrying uncertain labels (i.e., ``$-1$'' entries in the CheXpert-style label vocabulary) are dropped during harmonization, yielding 473{,}719 training and 52{,}408 validation images after patient-level splitting. Each dataset is instantiated with a single constructor call specifying only the image directory; metadata CSV paths, label mappings, and patient-level splits are handled automatically. The entire experiment requires no dataset-specific code: combining three heterogeneous datasets (JPEG, DICOM, and PNG formats with different metadata CSV schemas) is reduced to three constructor calls and a concatenation, as shown in Code~\ref{lst:multi-dataset}.

\begin{lstlisting}[caption={Assembling the multi-dataset training set used by RadHarmony-ViT. Three heterogeneous chest radiograph datasets (JPEG/DICOM/PNG) are loaded, harmonized, and concatenated with no dataset-specific code.},label={lst:multi-dataset},float=t]
from torch.utils.data import ConcatDataset
from radharmony.dataset import (
CheXpertDataset, MIMICCXRDataset, ChestXray14Dataset,
)

tfm = RadiologyTransform2D(img_size=256).get_transform()
datasets = [
    CheXpertDataset(base_image_dir=CHEXPERT_DIR, transform=tfm, drop_uncertain=True),
    MIMICCXRDataset(base_image_dir=MIMIC_DIR,    transform=tfm, drop_uncertain=True),
    ChestXray14Dataset(base_image_dir=CXR14_DIR, transform=tfm),  # already 0/1
]
train_sets, val_sets = zip(*(ds.get_datasets(n_splits=10) for ds in datasets))
train_set = ConcatDataset(train_sets) # 473,719 images
val_set = ConcatDataset(val_sets) # 52,408 images
\end{lstlisting}

To showcase the framework's flexibility, we additionally train two variants of RadHarmony-ViT. First, a CheXpert-only variant pretrained on CheXpert alone, holding all other hyperparameters fixed, which isolates the effect of including additional harmonized datasets from model and optimization choices. Second, a 512-resolution variant obtained by continuing pretraining of the full multi-dataset model at input resolution 512 for 5 additional epochs at a lower learning rate, exploiting NaFlex's native support for variable-resolution inputs to demonstrate that RadHarmony-ViT scales to higher resolutions without architectural changes. Both variants reuse the same RadHarmony data-loading code, differing only in the datasets and the image sizes, as illustrated in Code~\ref{lst:variants}.

\begin{lstlisting}[caption={Training the CheXpert-only and 512-resolution variants. Switching datasets or resolution requires changing only the constructor list or the \texttt{img\_size} parameter; all other code remains identical to Code~\ref{lst:multi-dataset}.},label={lst:variants},float=t]
# Variant 1: CheXpert-only (single dataset, same resolution)
tfm = RadiologyTransform2D(img_size=256).get_transform()
datasets = [CheXpertDataset(base_image_dir=CHEXPERT_DIR, transform=tfm)]

# Variant 2: Full model at 512 resolution (same datasets, higher resolution)
tfm_512 = RadiologyTransform2D(img_size=512).get_transform()
datasets = [
    CheXpertDataset(base_image_dir=CHEXPERT_DIR, transform=tfm_512),
    MIMICCXRDataset(base_image_dir=MIMIC_DIR,    transform=tfm_512),
    ChestXray14Dataset(base_image_dir=CXR14_DIR, transform=tfm_512),
]
\end{lstlisting}

We evaluate the pretrained backbones on VinDr-CXR~\cite{ref11} as a held-out benchmark. Unlike the NLP-derived labels used in the pretraining datasets, VinDr-CXR carries high-quality radiologist-drawn bounding boxes, making it a particularly reliable target for downstream evaluation. We train a linear probe (logistic regression on frozen embeddings), as outlined in Code~\ref{lst:eval}. VinDr-CXR ships with an official 15{,}000-image training split and a 3{,}000-image 3-radiologist consensus test split; we concatenate both splits into a single 18{,}000-image pool and run 5-fold cross-validation over the combined pool. Within each fold we additionally sweep the number of training samples $n \in \{1500, 3750, 7500, 11250, 14400\}$, uniformly sub-sampling from the fold's training partition (the largest size corresponds to using essentially all available fold-training examples). We select the same seven focal findings as RAD-DINO~\cite{ref_raddino} (lung opacity, cardiomegaly, pleural thickening, aortic enlargement, pulmonary fibrosis, tuberculosis, pleural effusion) and report per-label and macro-averaged AUROC/AUPRC (mean $\pm$ standard deviation across folds).

\begin{lstlisting}[caption={Linear probe evaluation on VinDr-CXR. The held-out dataset is loaded through RadHarmony with the same API used for pretraining; \texttt{get\_folds} yields patient-level 5-fold splits. Embeddings are extracted from the frozen backbone and evaluated with per-label logistic regression.},label={lst:eval},float=t]
from radharmony.dataset import VinDrCXRTrainDataset, RadiologyTransform2D
from sklearn.linear_model import LogisticRegression

tfm = RadiologyTransform2D(img_size=256, output_keys={"img", "cls"}).get_transform()
ds = VinDrCXRTrainDataset(base_image_dir=VINDR_DIR, transform=tfm, output_cls=True)

for fold, (train_ds, val_ds) in enumerate(ds.get_folds(n_splits=5)):
    train_feats, train_labels = extract_embeddings(model, DataLoader(train_ds))
    val_feats,   val_labels   = extract_embeddings(model, DataLoader(val_ds))
    for label_i in range(train_labels.shape[1]):
        clf = LogisticRegression(max_iter=5000)
        clf.fit(train_feats, train_labels[:, label_i])
        probs = clf.predict_proba(val_feats)[:, 1]
\end{lstlisting}

\section{Supported datasets}\label{sec:datasets}

RadHarmony currently supports 24 radiological datasets spanning chest radiographs, computed tomography, and magnetic resonance imaging of the spine, together covering the abdomen, cervical spine, lumbar spine, and paediatric hand (bone-age) alongside chest imaging (Table~\ref{tab:datasets}). The primary focus of the current release is chest imaging (17 of the 24 datasets are chest radiographs), which is also the most thoroughly tested and documented subset of the framework. The included CT and MRI datasets serve as early demonstrations of the framework's modality-agnostic design (marked as beta in Table~\ref{tab:datasets}), and broader coverage of these and other modalities remains future work. The datasets span five image formats (DICOM, JPEG, PNG, NIfTI, NumPy) and four annotation types (classification labels, segmentation masks, bounding boxes, radiology reports). Table~\ref{tab:finding-labels} summarizes how the finding-label vocabularies overlap across the classical chest-imaging datasets, highlighting the core set of findings that multiple datasets share and that cross-dataset training and evaluation can leverage.

\begin{table}[!htbp]
  \centering
  \caption{Datasets supported by RadHarmony. Rows marked $\dagger$ have beta support and are still under active testing. Formats listed with a slash (e.g.\ DICOM/PNG) are available in either format at construction time.}\label{tab:datasets}
  \footnotesize
  \begin{tabular}{@{}lllll@{}}
    \toprule
    \textbf{Dataset} & \textbf{Modality} & \textbf{Format} & \textbf{Images} & \textbf{Annotations} \\
    \midrule
    CheXpert~\cite{ref1}                     & CXR        & JPEG       & 224k   & 14 findings \\
    CheXpert-Plus~\cite{ref8}                & CXR        & DICOM      & 224k   & 14 findings + reports \\
    ChestX-ray14~\cite{ref3}                 & CXR        & PNG        & 112k   & 15 findings + bounding boxes \\
    MIMIC-CXR~\cite{ref2}                    & CXR        & DICOM      & 377k   & 14 findings + reports \\
    MIMIC-CXR-JPG~\cite{ref9}                & CXR        & JPEG       & 377k   & 14 findings \\
    PadChest~\cite{padchest}                  & CXR        & PNG        & 161k   & 193 findings \\
    BRAX~\cite{brax}                          & CXR        & DICOM/PNG  & 41k    & 14 findings \\
    ReXGradient-160K~\cite{rexgradient}       & CXR        & PNG        & 273k   & Radiology reports \\
    VinDr-CXR~\cite{ref11}                    & CXR        & DICOM      & 18k    & 28 findings + bounding boxes \\
    RSNA Pneumonia~\cite{ref17}               & CXR        & DICOM      & 27k    & 3 classes + bounding boxes \\
    SIIM-ACR PTX~\cite{ref10}                 & CXR        & DICOM      & 13k    & Pneumothorax masks \\
    SIIM COVID-19~\cite{siim_covid19}         & CXR        & DICOM      & 6.3k   & 4 appearance classes + bounding boxes \\
    RANZCR CLiP~\cite{ranzcr_clip}            & CXR        & JPEG       & 30k    & 11 catheter/line position labels \\
    TAIX-Ray~\cite{taix_ray}                  & CXR        & PNG        & 215k   & 8 findings (binary/ordinal) \\
    OpenI IU CXR~\cite{openi_iu}              & CXR        & PNG/DICOM  & 7.5k   & 8 findings + reports \\
    Montgomery County CXR~\cite{jaeger2014}   & CXR        & PNG        & 138    & TB classification + lung masks \\
    Shenzhen Hospital CXR~\cite{jaeger2014}   & CXR        & PNG        & 662    & TB classification \\
    RSNA Pediatric Bone Age$^\dagger$~\cite{rsna_bone_age}          & Radiograph & PNG        & 12.6k  & Age regression \\
    CT-RATE$^\dagger$~\cite{ref4}             & CT         & NIfTI      & 50k    & 18 findings \\
    RAD-ChestCT$^\dagger$~\cite{ref12}        & CT         & NumPy      & 36k    & 84 findings + lung bounding boxes \\
    RSNA PE Detection$^\dagger$~\cite{rsna_pe}                       & CT         & DICOM      & 7.3k   & 13 PE labels \\
    RSNA 2022 Cervical Spine$^\dagger$~\cite{rsna_cervical_2022}     & CT         & DICOM      & 2k     & 8 fracture labels + masks/bboxes \\
    RSNA 2023 Abdominal Trauma$^\dagger$~\cite{rsna_abdominal_2023}  & CT         & DICOM      & 4.7k   & 14 injury labels \\
    RSNA 2024 Lumbar Spine$^\dagger$~\cite{rsna_lumbar_2024}         & MRI        & DICOM      & 6.3k   & 75 severity labels + bounding boxes \\
    \botrule
  \end{tabular}
\end{table}

\begin{table}[!htbp]
  \centering
  \caption{Finding labels across the supported chest radiograph and CT datasets. Columns group label-compatible datasets: \textbf{ChX-fam} = CheXpert, CheXpert-Plus, MIMIC-CXR, MIMIC-CXR-JPG (shared 14-label schema); \textbf{NIH} = ChestX-ray14 (and the bbox subset); \textbf{VinDr} = VinDr-CXR; \textbf{RSNA} = RSNA Pneumonia; \textbf{SIIM} = SIIM-ACR Pneumothorax; \textbf{CT-R} = CT-RATE; \textbf{RCT} = RAD-ChestCT. Rows list every finding that appears in at least two dataset groups; shaded rows mark findings present in at least four groups. Newer chest-radiograph datasets (BRAX, PadChest, ReXGradient-160K, OpenI IU CXR, Montgomery, Shenzhen, RANZCR CLiP, TAIX-Ray) mostly reuse the ChX-fam 14-label schema or use dataset-specific labels documented in the online documentation (\url{https://f10409.github.io/RadHarmony/}), so no additional columns are shown here.}\label{tab:finding-labels}
  \footnotesize
  \begin{tabular}{@{}lccccccc@{}}
    \toprule
    \textbf{Finding} & \textbf{ChX-fam} & \textbf{NIH} & \textbf{VinDr} & \textbf{RSNA} & \textbf{SIIM} & \textbf{CT-R} & \textbf{RCT} \\
    \midrule
    \rowcolor{gray!15} Atelectasis        & \yes & \yes & \yes & \no  & \no  & \yes & \yes \\
    \rowcolor{gray!15} Cardiomegaly       & \yes & \yes & \yes & \no  & \no  & \yes & \yes \\
    \rowcolor{gray!15} Consolidation      & \yes & \yes & \yes & \no  & \no  & \yes & \yes \\
    \rowcolor{gray!15} Pleural effusion   & \yes & \yes\footnotemark[1] & \yes & \no  & \no  & \yes & \yes \\
    \rowcolor{gray!15} Lung opacity       & \yes & \no  & \yes & \yes & \no  & \yes & \yes \\
    \rowcolor{gray!15} Pneumothorax       & \yes & \yes & \yes & \no  & \yes & \no  & \yes \\
    \rowcolor{gray!15} Pneumonia          & \yes & \yes & \yes & \no  & \no  & \no  & \yes \\
    \rowcolor{gray!15} Edema              & \yes & \yes & \yes & \no  & \no  & \no  & \yes \\
    \rowcolor{gray!15} Emphysema          & \no  & \yes & \yes & \no  & \no  & \yes & \yes \\
    \rowcolor{gray!15} Nodule/Mass        & \no  & \yes & \yes & \no  & \no  & \yes & \yes \\
    \rowcolor{gray!15} Fibrosis           & \no  & \yes & \yes & \no  & \no  & \yes\footnotemark[2] & \yes \\
    Pleural thickening & \no  & \yes & \yes & \no  & \no  & \no  & \yes \\
    Fracture           & \yes & \no  & \yes & \no  & \no  & \no  & \yes \\
    Lung lesion        & \yes & \no  & \yes & \no  & \no  & \no  & \yes \\
    Calcification      & \no  & \no  & \yes & \no  & \no  & \yes\footnotemark[3] & \yes \\
    Hernia             & \no  & \yes & \no  & \no  & \no  & \yes\footnotemark[4] & \yes \\
    Infiltration       & \no  & \yes & \yes & \no  & \no  & \no  & \yes\footnotemark[5] \\
    Tuberculosis       & \no  & \no  & \yes & \no  & \no  & \no  & \yes \\
    Pericardial effusion & \no  & \no  & \no  & \no  & \no  & \yes & \yes \\
    \botrule
  \end{tabular}
  \footnotetext[1]{ChestX-ray14 labels this as ``Effusion''.}
  \footnotetext[2]{CT-RATE labels this as ``Pulmonary fibrotic sequela''.}
  \footnotetext[3]{CT-RATE distinguishes arterial-wall and coronary-artery-wall calcification.}
  \footnotetext[4]{CT-RATE labels ``Hiatal hernia''.}
  \footnotetext[5]{RAD-ChestCT labels this as ``infiltrate''.}
\end{table}

\section{Results}\label{sec:results}

\noindent\textbf{Effect of additional harmonized datasets.} Table~\ref{tab:results} compares the full multi-dataset RadHarmony-ViT against the CheXpert-only ablation. Both models share identical architecture, optimizer, and training budget; the only difference is whether the harmonizer concatenates MIMIC-CXR and ChestX-ray14 alongside CheXpert. At full label budget ($n{=}14{,}400$), the two models perform comparably (macro AUROC 0.905 vs.\ 0.903; macro AUPRC 0.483 vs.\ 0.478), with neither consistently dominating across findings. We frame this experiment as a capability demonstration rather than a performance claim: obtaining a rigorous null of this kind ordinarily requires substantial per-dataset engineering, whereas RadHarmony reduces the setup to the constructor list shown in Code~\ref{lst:multi-dataset}.

\begin{table}[!htbp]
  \centering
  \caption{Per-finding VinDr-CXR linear-probe AUROC at $n{=}14{,}400$ training samples per fold (5-fold CV on the combined 18k pool). Columns are the three RadHarmony-ViT variants. Label abbreviations: LO (lung opacity), CM (cardiomegaly), PL-T (pleural thickening), AE (aortic enlargement), PF (pulmonary fibrosis), TB (tuberculosis), PE (pleural effusion); Agg.\ is the macro average. Full = CheXpert + MIMIC-CXR + ChestX-ray14.}\label{tab:results}
  \begin{tabular}{@{}lccc@{}}
    \toprule
    & \textbf{CheXpert-only} & \textbf{Full} & \textbf{Full} \\
    \textbf{Finding} & 256 & 256 & 512 \\
    \midrule
    LO   & $0.864 \pm 0.011$ & $0.878 \pm 0.010$ & $0.877 \pm 0.013$ \\
    CM   & $0.954 \pm 0.005$ & $0.955 \pm 0.004$ & $0.961 \pm 0.002$ \\
    PL-T & $0.886 \pm 0.008$ & $0.875 \pm 0.006$ & $0.893 \pm 0.008$ \\
    AE   & $0.948 \pm 0.003$ & $0.952 \pm 0.002$ & $0.956 \pm 0.003$ \\
    PF   & $0.874 \pm 0.003$ & $0.866 \pm 0.011$ & $0.872 \pm 0.009$ \\
    TB   & $0.879 \pm 0.014$ & $0.870 \pm 0.011$ & $0.873 \pm 0.007$ \\
    PE   & $0.929 \pm 0.007$ & $0.925 \pm 0.011$ & $0.931 \pm 0.009$ \\
    \midrule
    Agg. & $0.905 \pm 0.035$ & $0.903 \pm 0.037$ & $0.909 \pm 0.036$ \\
    \botrule
  \end{tabular}
\end{table}

\begin{table}[!htbp]
  \centering
  \caption{Per-finding VinDr-CXR linear-probe AUPRC at $n{=}14{,}400$ training samples per fold (5-fold CV on the combined 18k pool). Columns are the three RadHarmony-ViT variants. Label abbreviations follow Table~\ref{tab:results}. Full = CheXpert + MIMIC-CXR + ChestX-ray14.}\label{tab:results-auprc}
  \begin{tabular}{@{}lccc@{}}
    \toprule
    & \textbf{CheXpert-only} & \textbf{Full} & \textbf{Full} \\
    \textbf{Finding} & 256 & 256 & 512 \\
    \midrule
    LO   & $0.263 \pm 0.020$ & $0.284 \pm 0.032$ & $0.279 \pm 0.034$ \\
    CM   & $0.739 \pm 0.022$ & $0.746 \pm 0.017$ & $0.776 \pm 0.018$ \\
    PL-T & $0.364 \pm 0.041$ & $0.319 \pm 0.027$ & $0.352 \pm 0.032$ \\
    AE   & $0.737 \pm 0.028$ & $0.759 \pm 0.026$ & $0.768 \pm 0.018$ \\
    PF   & $0.369 \pm 0.024$ & $0.368 \pm 0.009$ & $0.392 \pm 0.024$ \\
    TB   & $0.334 \pm 0.045$ & $0.312 \pm 0.013$ & $0.332 \pm 0.015$ \\
    PE   & $0.573 \pm 0.041$ & $0.559 \pm 0.050$ & $0.591 \pm 0.055$ \\
    \midrule
    Agg. & $0.483 \pm 0.184$ & $0.478 \pm 0.193$ & $0.499 \pm 0.195$ \\
    \botrule
  \end{tabular}
\end{table}

\begin{figure}[!htbp]
  \centering
  \includegraphics[width=0.48\linewidth]{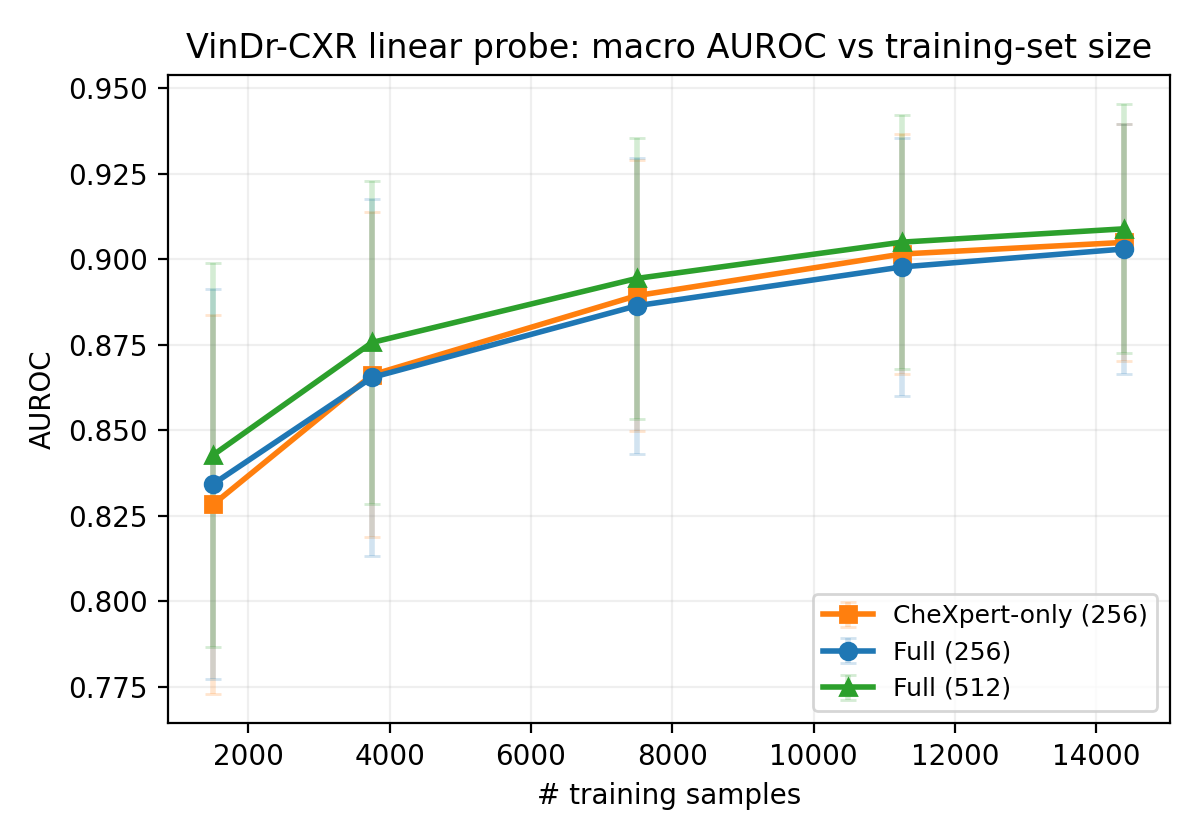}\hfill
  \includegraphics[width=0.48\linewidth]{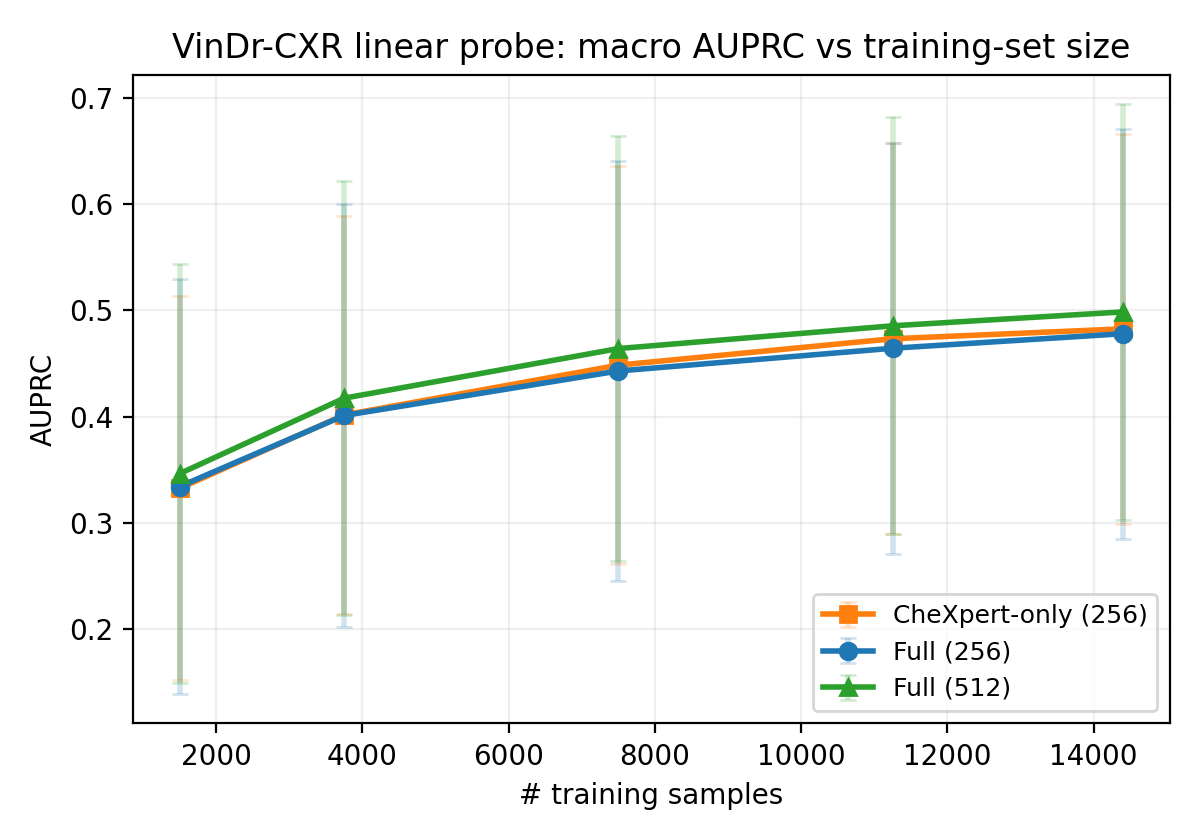}
  \caption{VinDr-CXR linear-probe macro performance vs number of training samples for the three RadHarmony-ViT variants (CheXpert-only, Full, Full@512). Left: macro AUROC; right: macro AUPRC. Each curve is the macro average over the seven focal findings; error bars show standard deviation across 5 cross-validation folds.}\label{fig:vindr-sweep}
\end{figure}

\noindent\textbf{Linear-probe data scaling.} Figure~\ref{fig:vindr-sweep} reports the macro-averaged VinDr-CXR linear-probe performance as a function of training-set size $n \in \{1500, 3750, 7500, 11250, 14400\}$ for all three RadHarmony-ViT variants. Significance across the 5 cross-validation folds is assessed with two-sided paired Student's $t$-tests on per-fold macro-averaged metrics. The full multi-dataset model and the CheXpert-only variant remain close across all label budgets, and the small gap at $n{=}1{,}500$ falls within paired-fold noise ($\Delta$ AUROC $+0.006$, $p{=}0.26$) and reverses sign at $n{=}3{,}750$ (Table~\ref{tab:paired-stats}). In contrast, the Full(512) versus Full(256) comparison yields consistently positive, statistically significant AUPRC differences at the same low label budgets ($\Delta$ AUPRC $+0.012$, $p{=}0.043$ at $n{=}1{,}500$; $+0.016$, $p{=}0.032$ at $n{=}3{,}750$), indicating a small but real resolution effect. Regardless of the sign or magnitude of any individual outcome, RadHarmony's unified dataset interface makes it straightforward to implement such evaluation protocols, including pool construction, fold management, and subsampling, without dataset-specific glue code.

\begin{table}[!htbp]
  \centering
  \caption{Paired-fold statistics at low label budgets ($n{=}1{,}500$ and $n{=}3{,}750$) for the diversity comparison (Full vs CheXpert-only, both at input resolution 256) and the resolution comparison (Full at 512 vs 256). $\Delta$ is the paired mean difference across 5 cross-validation folds; $d_z$ is the standardized paired effect size; $p$ is from a two-sided paired Student's $t$-test across the 5 folds. 95\% confidence intervals in brackets.}\label{tab:paired-stats}
  \footnotesize
  \begin{tabular}{@{}llcccc@{}}
    \toprule
    \textbf{Comparison} & \textbf{Metric} & \textbf{$n$} & \textbf{$\Delta$ (95\% CI)} & \textbf{$d_z$ (95\% CI)} & \textbf{$p$} \\
    \midrule
    Full(256) vs CheX(256) & AUROC & 1{,}500 & $+0.006\ [-0.007,\ +0.019]$ & $+0.59\ [-0.40,\ +1.52]$   & $0.26$ \\
    Full(256) vs CheX(256) & AUPRC & 1{,}500 & $+0.002\ [-0.013,\ +0.016]$ & $+0.14\ [-0.75,\ +1.01]$   & $0.77$ \\
    Full(256) vs CheX(256) & AUROC & 3{,}750 & $-0.001\ [-0.004,\ +0.003]$ & $-0.27\ [-1.15,\ +0.64]$   & $0.58$ \\
    Full(256) vs CheX(256) & AUPRC & 3{,}750 & $-0.001\ [-0.017,\ +0.016]$ & $-0.04\ [-0.91,\ +0.84]$   & $0.94$ \\
    \midrule
    Full(512) vs Full(256) & AUROC & 1{,}500 & $+0.009\ [-0.002,\ +0.019]$ & $+1.05\ [-0.11,\ +2.13]$   & $0.079$ \\
    Full(512) vs Full(256) & AUPRC & 1{,}500 & $+0.012\ [+0.001,\ +0.024]$ & $+1.31\ [+0.03,\ +2.51]$   & $0.043$ \\
    Full(512) vs Full(256) & AUROC & 3{,}750 & $+0.010\ [+0.009,\ +0.011]$ & $+12.48\ [+4.28,\ +20.86]$ & $<\!0.001$ \\
    Full(512) vs Full(256) & AUPRC & 3{,}750 & $+0.016\ [+0.002,\ +0.030]$ & $+1.45\ [+0.11,\ +2.72]$   & $0.032$ \\
    \botrule
  \end{tabular}
\end{table}

\noindent\textbf{Resolution scaling.} Continuing pretraining at input resolution 512 for 5 additional epochs marginally improves macro AUROC on VinDr-CXR from 0.903 to 0.909 (+0.6\%) and macro AUPRC from 0.478 to 0.499 (+2.1\%) without modifying the backbone, confirming that RadHarmony-ViT inherits NaFlex's variable-resolution capability and can be scaled by users with modest additional compute. For reference, RAD-DINO~\cite{ref_raddino} reports a macro AUPRC of 0.528 on the same seven findings, although their cross-validation protocol is not fully documented and therefore the numbers are not strictly comparable.

\section{Discussion}\label{sec:discussion}

RadHarmony-ViT serves as a concrete demonstration that the framework removes the engineering friction typically associated with multi-dataset pretraining in medical imaging. The CheXpert-only ablation is a capability demonstration rather than a performance claim: it isolates dataset diversity by holding every other knob fixed, and the resulting null at full label budget (with the small $n{=}1{,}500$ gap reversing sign at $n{=}3{,}750$) is itself a rigorous outcome that a unified data layer makes inexpensive to obtain. In practice, adding MIMIC-CXR and ChestX-ray14 to a CheXpert-only run requires changing a single list of dataset constructors, no per-dataset preprocessing, label-mapping, or split-handling code. The 512-resolution variant similarly shows that scaling input resolution is a one-argument change made possible by NaFlex, with the harmonizer automatically handling the change for all underlying datasets. More broadly, RadHarmony's unified data interface together with the NaFlex backbone's flexibility makes RadHarmony-ViT easy to adapt to other downstream tasks and datasets beyond the linear-probe setting explored here. Together, these results support the broader claim that a unified data layer is crucial for the kind of rapid, low-friction experimentation that medical imaging research increasingly demands, whether the outcome of a given experiment is positive or null.

\noindent\textbf{Limitations.}
The current release centers on chest imaging (in particular chest radiographs), which is the most thoroughly tested and documented subset of the framework; support for CT and MRI is provided as an early demonstration of the modality-agnostic design and is still under active testing. The AI-assisted integration workflow is an assistive tool, not an autonomous system: end-to-end correctness relies on the reviewing user, and broader qualitative evaluation across additional datasets, users, and annotation types remains future work. RadHarmony-ViT is a reference baseline intended to illustrate the framework rather than a model optimized for downstream performance, and integrating new datasets may still require manual effort when their metadata or annotation formats fall outside patterns already covered by the framework.

\noindent\textbf{Future work.}
An immediate next step is to graduate the current CT and MRI support from beta by bringing their 3-D preprocessing and augmentation pipelines to the same maturity as the chest radiograph pipeline. Beyond that, we plan to extend coverage to further anatomical regions (e.g., mammography, musculoskeletal radiographs) and modalities (e.g., ultrasound, PET, digital pathology). Each of these introduces harmonization challenges (multi-view acquisitions, modality-specific preprocessing, richer study-level hierarchies) that will likely require extending the harmonized schema and introducing modality-specific preprocessing steps. We view these as promising directions for community contribution, aided by the AI-assisted integration workflow introduced in this paper. A complementary direction is to extend the current AI agent skill into a more agentic workflow in which the agent inspects candidate dataset metadata, drafts a harmonizer, runs integrity checks, and opens a pull request for human review, moving from AI-assisted integration toward a semi-automatic harmonization pipeline with the human kept in the loop for final validation. Finally, as the project matures into a broader community effort, formalizing more comprehensive governance rules (contribution guidelines, code review standards, release management, and deprecation policies) will be needed to complement the current CI-based quality gate.

\backmatter

\bmhead{Acknowledgements}
The authors thank Beatrice Brown-Mulry, Rohan Satya Isaac, Aawez Mansuri, Chiratidzo Sanyika, and Anjana Dissanayaka for helpful discussions and feedback throughout the development of this work.

\section*{Declarations}

\begin{itemize}
\item \textbf{Funding.} This study was supported by the AI Image Extraction Core, an Emory Integrated Core Facility. Dr.\ Gichoya is a 2022 Robert Wood Johnson Foundation Harold Amos Medical Faculty Development Program and declares support from Lacuna Fund (\#67), Gordon and Betty Moore Foundation, NIH (NIBIB) MIDRC grant under contracts 75N92020C00008 and 75N92020C00021, NHLBI Award Number R01HL167811, and NIH common fund award 1R25OD039834-01.
\item \textbf{Conflict of interest.} The authors declare no competing interests.
\item \textbf{Ethics approval and consent to participate.} Not applicable; this work uses publicly released datasets only.
\item \textbf{Data availability.} RadHarmony does not redistribute or host any of the underlying datasets. Users obtain each dataset directly from its original publisher under that dataset's own license and terms. Per-dataset download instructions, expected directory layouts, and label documentation are provided in the online documentation at \url{https://f10409.github.io/RadHarmony/}.
\item \textbf{Code availability.} All code, including harmonizers, dataset classes, transforms, the interactive visualizer, and the AI-assisted integration skill, is released at \url{https://github.com/f10409/RadHarmony}. Pretrained RadHarmony-ViT model weights are released through the same repository.
\item \textbf{Author contribution.} All authors made a significant contribution to the work reported, whether that is in the conception, study design, execution, acquisition of data, analysis and interpretation, or in all these areas; took part in drafting, revising or critically reviewing the article; gave final approval of the version to be published; have agreed on the journal to which the article has been submitted; and agree to be accountable for all aspects of the work.
\item \textbf{Declaration of generative AI and AI-assisted technologies.} During the preparation of this work, the authors used Claude Code to assist with language editing, grammar correction, and the summarization of tabular results. After using this tool, the authors reviewed and edited the content as needed and take full responsibility for the content of the published article.
\end{itemize}

\begin{appendices}

\section{Dataset configuration schema}\label{sec:appendix-config}

The AI-assisted dataset integration workflow (Section~\ref{sec:ai-integration}) is anchored by a declarative JSON configuration file. This file describes how to read and interpret a dataset: where the files are stored, how to extract patient and study identifiers, which columns contain labels, and what annotation types are available. Code~\ref{lst:rsna-config} shows the complete configuration used to integrate the RSNA Pneumonia dataset.

Each top-level key maps to one aspect of the harmonization: \texttt{primary\_csv} locates the raw metadata source, \texttt{patient\_id} and \texttt{study\_id} specify how to derive identifiers, \texttt{image\_path} describes the file naming convention, \texttt{labels} enumerates the label columns and their uncertain-value handling, and \texttt{bboxes}/\texttt{masks}/\texttt{reports} declare which annotation types are available. The \texttt{\_comment} fields serve as inline documentation for the AI agent, providing dataset-specific context that guides code generation. Fields set to \texttt{null} indicate that the information is either not applicable or must be derived programmatically (e.g., constructing image paths from DICOM UIDs rather than reading them from a CSV column).

\begingroup
\lstset{style=jsonstyle}
\begin{lstlisting}[caption={Complete JSON configuration for the RSNA Pneumonia dataset. This file drives the AI-assisted integration workflow: the AI agent reads it to generate the harmonizer, dataset class, and integrity checks. Comment fields provide context that guides the agent's code generation decisions.},label={lst:rsna-config}]
{
  "_comment": "RSNA Pneumonia Detection Challenge (2018).
    30,000 frontal chest X-rays in DICOM format,
    a subset of NIH ChestX-ray14.",

  "dataset_name": "RSNA Pneumonia",
  "registry_key": "rsna_pneumonia",
  "modality": "2D",
  "image_format": "DICOM",
  "base_image_dir": "~/Downloads/rsna",

  "primary_csv": {
    "_comment": "No CSV -- annotations come from an
      MD.ai JSON export.",
    "path": null,
    "filename_variants": [
      "pneumonia-challenge-annotations-adjudicated
       -kaggle_2018.json"
    ]
  },

  "patient_id": {
    "column": null,
    "derivation": "Extract StudyInstanceUID
      from each annotation"
  },

  "study_id": {
    "column": null,
    "derivation": "Same as StudyInstanceUID"
  },

  "image_path": {
    "column": null,
    "derivation": "Construct as StudyInstanceUID/
      SeriesInstanceUID/SOPInstanceUID.dcm",
    "prefix_to_strip": null
  },

  "labels": {
    "_comment": "Three classes from the 'Calculated'
      label group: Normal, No Lung Opacity / Not Normal,
      Lung Opacity. One-hot encoded.",
    "columns": [
      "Lung Opacity",
      "No Lung Opacity / Not Normal",
      "Normal"
    ],
    "separate_csv": {
      "path": null,
      "filename_variants": [],
      "join_columns": []
    },
    "uncertain_handling": {
      "nan_action": "to_zero",
      "negative_one_action": null,
      "drop_uncertain_rows": false
    }
  },

  "view_position": {
    "_comment": "All images are frontal chest X-rays.",
    "column": null,
    "source": null
  },

  "masks": {
    "supported": false,
    "column": null,
    "format": null
  },

  "bboxes": {
    "_comment": "Bounding boxes for Lung Opacity
      annotations. Pixel coordinates on 1024x1024.",
    "supported": true,
    "column": "annotation.data:
      {x, y, width, height} in JSON"
  },

  "reports": {
    "supported": false,
    "format": null
  },

  "extra_init_params": {
    "json_path": "Path to the MD.ai JSON export file",
    "label_group": "Which label group to use.
      Default 'Calculated' for final consensus."
  },

  "hu_window": null,
  "custom_preprocessing": "Parse MD.ai JSON format.
    Extract annotations from the 'Calculated' group.
    Build per-image rows with one-hot class labels
    and aggregated bounding boxes."
}
\end{lstlisting}
\endgroup

\end{appendices}

\bibliography{ref}

@article{ref1,
   author = {Jeremy Irvin and Pranav Rajpurkar and Michael Ko and Yifan Yu and Silviana Ciurea-Ilcus and Chris Chute and Henrik Marklund and Behzad Haghgoo and Robyn Ball and Katie Shpanskaya and Jayne Seekins and David A. Mong and Safwan S. Halabi and Jesse K. Sandberg and Ricky Jones and David B. Larson and Curtis P. Langlotz and Bhavik N. Patel and Matthew P. Lungren and Andrew Y. Ng},
   month = {1},
   title = {CheXpert: A Large Chest Radiograph Dataset with Uncertainty Labels and Expert Comparison},
   journal = {Proceedings of the AAAI Conference on Artificial Intelligence},
   url = {http://arxiv.org/abs/1901.07031},
   year = {2019}
}

@article{ref2,
   author = {Alistair E.W. Johnson and Tom J. Pollard and Seth J. Berkowitz and Nathaniel R. Greenbaum and Matthew P. Lungren and Chih-ying Deng and Roger G. Mark and Steven Horng},
   doi = {10.1038/s41597-019-0322-0},
   issn = {20524463},
   issue = {1},
   journal = {Scientific Data},
   month = {12},
   pmid = {31831740},
   publisher = {Nature Research},
   title = {MIMIC-CXR, a de-identified publicly available database of chest radiographs with free-text reports},
   volume = {6},
   year = {2019}
}

@article{ref3,
   author = {Xiaosong Wang and Yifan Peng and Le Lu and Zhiyong Lu and Mohammadhadi Bagheri and Ronald M. Summers},
   doi = {10.1109/CVPR.2017.369},
   month = {12},
   title = {ChestX-ray8: Hospital-scale Chest X-ray Database and Benchmarks on Weakly-Supervised Classification and Localization of Common Thorax Diseases},
   journal = {Proceedings of the IEEE Conference on Computer Vision and Pattern Recognition (CVPR)},
   url = {http://arxiv.org/abs/1705.02315},
   year = {2017}
}

@article{ref4,
   author = {Ibrahim Ethem Hamamci and Sezgin Er and Chenyu Wang and Furkan Almas and Ayse Gulnihan Simsek and Sevval Nil Esirgun and Irem Doga and Omer Faruk Durugol and Weicheng Dai and Murong Xu and Muhammed Furkan Dasdelen and Bastian Wittmann and Tamaz Amiranashvili and Enis Simsar and Mehmet Simsar and Emine Bensu Erdemir and Abdullah Alanbay and Anjany Sekuboyina and Berkan Lafci and Christian Bluethgen and Kayhan Batmanghelich and Mehmet Kemal Ozdemir and Bjoern Menze},
   doi = {10.1038/s41551-025-01599-y},
   issn = {2157846X},
   journal = {Nature Biomedical Engineering},
   month = {2},
   title = {Developing Generalist Foundation Models from a Multimodal Dataset for 3D Computed Tomography},
   url = {http://arxiv.org/abs/2403.17834},
   year = {2025}
}

@article{ref5,
   author = {M. Jorge Cardoso and Wenqi Li and Richard Brown and Nic Ma and Eric Kerfoot and Yiheng Wang and Benjamin Murrey and Andriy Myronenko and Can Zhao and Dong Yang and Vishwesh Nath and Yufan He and Ziyue Xu and Ali Hatamizadeh and Wentao Zhu and Yun Liu and Mingxin Zheng and Yucheng Tang and Isaac Yang and Michael Zephyr and Behrooz Hashemian and Sachidanand Alle and Mohammad Zalbagi Darestani and Charlie Budd and Marc Modat and Tom Vercauteren and Guotai Wang and Yiwen Li and Yipeng Hu and Yunguan Fu and Benjamin Gorman and Hans Johnson and Brad Genereaux and Barbaros S. Erdal and Vikash Gupta and Andres Diaz-Pinto and Andre Dourson and Lena Maier-Hein and Paul F. Jaeger and Michael Baumgartner and Jayashree Kalpathy-Cramer and Mona Flores and Justin Kirby and Lee A. D. Cooper and Holger R. Roth and Daguang Xu and David Bericat and Ralf Floca and S. Kevin Zhou and Haris Shuaib and Keyvan Farahani and Klaus H. Maier-Hein and Stephen Aylward and Prerna Dogra and Sebastien Ourselin and Andrew Feng},
   month = {11},
   title = {MONAI: An open-source framework for deep learning in healthcare},
   url = {http://arxiv.org/abs/2211.02701},
   year = {2022}
}

@article{ref6,
   author = {Fernando Pérez-García and Rachel Sparks and Sébastien Ourselin},
   doi = {10.1016/j.cmpb.2021.106236},
   issn = {18727565},
   journal = {Computer Methods and Programs in Biomedicine},
   month = {9},
   pmid = {34311413},
   publisher = {Elsevier Ireland Ltd},
   title = {TorchIO: A Python library for efficient loading, preprocessing, augmentation and patch-based sampling of medical images in deep learning},
   volume = {208},
   year = {2021}
}

@article{ref7,
   author = {Jiancheng Yang and Rui Shi and Donglai Wei and Zequan Liu and Lin Zhao and Bilian Ke and Hanspeter Pfister and Bingbing Ni},
   doi = {10.1038/s41597-022-01721-8},
   journal = {Scientific Data},
   month = {9},
   title = {MedMNIST v2 -- A large-scale lightweight benchmark for 2D and 3D biomedical image classification},
   url = {http://arxiv.org/abs/2110.14795},
   volume = {10},
   issue = {1},
   year = {2023}
}

@article{ref8,
   author = {Pierre Chambon and Jean-Benoit Delbrouck and Thomas Sounack and Shih-Cheng Huang and Zhihong Chen and Maya Varma and Steven Q. H. Truong and Chu The Chuong and Curtis P. Langlotz},
   title = {CheXpert Plus: Augmenting a Large Chest X-ray Dataset with Text Radiology Reports, Patient Demographics and Additional Image Formats},
   journal = {arXiv preprint arXiv:2405.19538},
   year = {2024}
}

@article{ref9,
   author = {Alistair E. W. Johnson and Tom J. Pollard and Nathaniel R. Greenbaum and Matthew P. Lungren and Chih-ying Deng and Yifan Peng and Zhiyong Lu and Roger G. Mark and Seth J. Berkowitz and Steven Horng},
   title = {MIMIC-CXR-JPG, a large publicly available database of labeled chest radiographs},
   journal = {arXiv preprint arXiv:1901.07042},
   year = {2019}
}

@misc{ref10,
   author = {{Society for Imaging Informatics in Medicine}},
   title = {SIIM-ACR Pneumothorax Segmentation},
   howpublished = {Kaggle},
   year = {2019},
   url = {https://www.kaggle.com/c/siim-acr-pneumothorax-segmentation}
}

@article{ref11,
   author = {Ha Q. Nguyen and Khanh Lam and Linh T. Le and Hieu H. Pham and Dat Q. Tran and Dung B. Nguyen and Dung D. Le and Chi M. Pham and Hang T.T. Tong and Diep H. Dinh and Cuong D. Do and Luu T. Doan and Cuong N. Nguyen and Binh T. Nguyen and Que V. Nguyen and Au D. Hoang and Hien N. Phan and Anh T. Nguyen and Phuong H. Ho and Dat T. Ngo and Nghia T. Nguyen and Nhan T. Nguyen and Minh Dao and Van Vu},
   doi = {10.1038/s41597-022-01498-w},
   issn = {20524463},
   issue = {1},
   journal = {Scientific Data},
   month = {12},
   pmid = {35858929},
   publisher = {Nature Research},
   title = {VinDr-CXR: An open dataset of chest X-rays with radiologist's annotations},
   volume = {9},
   year = {2022}
}

@article{ref12,
   author = {Rachel Lea Draelos and David Dov and Maciej A. Mazurowski and Joseph Y. Lo and Ricardo Henao and Geoffrey D. Rubin and Lawrence Carin},
   title = {Machine-learning-based multiple abnormality prediction with large-scale chest computed tomography volumes},
   journal = {Medical Image Analysis},
   volume = {67},
   year = {2021},
   doi = {10.1016/j.media.2020.101857}
}

@article{ref14,
   author = {Abubakar Abid and Ali Abdalla and Ali Abid and Dawood Khan and Abdulrahman Alfozan and James Zou},
   month = {6},
   title = {Gradio: Hassle-Free Sharing and Testing of ML Models in the Wild},
   journal = {ICML 2019 Workshop on Human in the Loop Learning},
   url = {http://arxiv.org/abs/1906.02569},
   year = {2019}
}

@article{ref15,
   author = {Randall Balestriero and Yann LeCun},
   month = {11},
   title = {LeJEPA: Provable and Scalable Self-Supervised Learning Without the Heuristics},
   journal = {arXiv preprint arXiv:2511.08544},
   url = {http://arxiv.org/abs/2511.08544},
   year = {2025}
}

@article{ref16,
   author = {Joseph Paul Cohen and Joseph D Viviano and Paul Bertin and Paul Morrison and Parsa Torabian and Matteo Guarrera and Matthew P Lungren and Akshay Chaudhari and Rupert Brooks and Mohammad Hashir and Hadrien Bertrand},
   journal = {Proceedings of Machine Learning Research},
   pages = {231-249},
   title = {TorchXRayVision: A library of chest X-ray datasets and models},
   volume = {172},
   year = {2022}
}

@article{ref17,
   author = {George Shih and Carol C. Wu and Safwan S. Halabi and Marc D. Kohli and Luciano M. Prevedello and Tessa S. Cook and Arjun Sharma and Judith K. Amorosa and Veronica Arteaga and Maya Galperin-Aizenberg and Ritu R. Gill and Myrna C.B. Godoy and Stephen Hobbs and Jean Jeudy and Archana Laroia and Palmi N. Shah and Dharshan Vummidi and Kavitha Yaddanapudi and Anouk Stein},
   doi = {10.1148/ryai.2019180041},
   issn = {26386100},
   issue = {1},
   journal = {Radiology: Artificial Intelligence},
   month = {1},
   publisher = {Radiological Society of North America Inc.},
   title = {Augmenting the national institutes of health chest radiograph dataset with expert annotations of possible pneumonia},
   volume = {1},
   year = {2019}
}

@article{ref18,
   author = {Michael Tschannen and Alexey Gritsenko and Xiao Wang and Muhammad Ferjad Naeem and Ibrahim Alabdulmohsin and Nikhil Parthasarathy and Talfan Evans and Lucas Beyer and Ye Xia and Basil Mustafa and Olivier Hénaff and Jeremiah Harmsen and Andreas Steiner and Xiaohua Zhai},
   month = {2},
   title = {SigLIP 2: Multilingual Vision-Language Encoders with Improved Semantic Understanding, Localization, and Dense Features},
   journal = {arXiv preprint arXiv:2502.14786},
   url = {http://arxiv.org/abs/2502.14786},
   year = {2025}
}

@article{ref_raddino,
   author = {Fernando Pérez-García and Harshita Sharma and Sam Bond-Taylor and Kenza Bouzid and Valentina Salvatelli and Maximilian Ilse and Shruthi Bannur and Daniel C. Castro and Anton Schwaighofer and Matthew P. Lungren and Maria Teodora Wetscherek and Noel Codella and Stephanie L. Hyland and Javier Alvarez-Valle and Ozan Oktay},
   doi = {10.1038/s42256-024-00965-w},
   issn = {25225839},
   issue = {1},
   journal = {Nature Machine Intelligence},
   month = {1},
   pages = {119-130},
   publisher = {Nature Research},
   title = {Exploring scalable medical image encoders beyond text supervision},
   volume = {7},
   year = {2025}
}

@article{Dapamede2025,
   author = {Theo Dapamede and Frank Li and Bardia Khosravi and Saptarshi Purkayastha and Hari Trivedi and Judy Gichoya},
   doi = {10.1007/s10278-025-01418-5},
   issn = {29482933},
   issue = {5},
   journal = {Journal of Imaging Informatics in Medicine},
   month = {10},
   pages = {3040-3048},
   publisher = {Springer Nature},
   title = {DICOM LUT is a Key Step in Medical Image Preprocessing Towards AI Generalizability},
   volume = {38},
   year = {2025}
}

@article{padchest,
   author = {Aurelia Bustos and Antonio Pertusa and Jose-Maria Salinas and Maria De La Iglesia-Vaya},
   title = {PadChest: A large chest x-ray image dataset with multi-label annotated reports},
   journal = {Medical Image Analysis},
   volume = {66},
   pages = {101797},
   year = {2020},
   month = {12},
   doi = {10.1016/j.media.2020.101797}
}

@article{brax,
   author = {Eduardo P. Reis and Joselisa P. De Paiva and Maria C. Da Silva and Guilherme A. Ribeiro and Victor F. Paiva and Lucas Bulgarelli and Hyun M. Lee and Paulo V. Santos and Vanessa M. Brito and Lucas T. Amaral and Gustavo L. Beraldo},
   title = {BRAX, {B}razilian labeled chest x-ray dataset},
   journal = {Scientific Data},
   volume = {9},
   number = {1},
   pages = {487},
   year = {2022},
   month = {8},
   doi = {10.1038/s41597-022-01608-8}
}

@article{rexgradient,
   author = {Xiaoman Zhang and Julián N. Acosta and Josh Miller and Ouwen Huang and Pranav Rajpurkar},
   title = {{ReXGradient-160K}: A large-scale publicly available dataset of chest radiographs with free-text reports},
   journal = {arXiv preprint arXiv:2505.00228},
   year = {2025},
   month = {5}
}

@misc{siim_covid19,
   author = {Andrew Kemp and Anna Zawacki and Chris Carr and George Shih and John Mongan and Julia Elliott and Kaiwen and Paras Lakhani and Phil Culliton},
   title = {{SIIM}-{FISABIO}-{RSNA} {COVID}-19 Detection},
   year = {2021},
   publisher = {Kaggle},
   howpublished = {\url{https://kaggle.com/competitions/siim-covid19-detection}}
}

@misc{ranzcr_clip,
   author = {Jarrel Seah and Jen and Maggie and Meng Law and Phil Culliton and Sarah Dowd},
   title = {{RANZCR CLiP} -- Catheter and Line Position Challenge},
   year = {2020},
   publisher = {Kaggle},
   howpublished = {\url{https://kaggle.com/competitions/ranzcr-clip-catheter-line-classification}}
}

@article{taix_ray,
   author = {Daniel Truhn and Daniel Geiger and Robert Siepmann and Maike S. von der Stück and Keno K. Bressem and Jakob N. Kather and Christiane Kuhl and Gustav Müller-Franzes and Sven Nebelung},
   title = {A comprehensive bedside chest radiography dataset with structured, itemized and graded radiologic reports},
   journal = {Scientific Data},
   volume = {13},
   number = {1},
   pages = {632},
   year = {2026},
   month = {4},
   doi = {10.1038/s41597-026-07271-7}
}

@article{openi_iu,
   author = {Dina Demner-Fushman and Marc D. Kohli and Marc B. Rosenman and Sonya E. Shooshan and Laritza Rodriguez and Sameer Antani and George R. Thoma and Clement J. McDonald},
   title = {Preparing a collection of radiology examinations for distribution and retrieval},
   journal = {Journal of the American Medical Informatics Association},
   volume = {23},
   number = {2},
   pages = {304-310},
   year = {2016},
   month = {3},
   doi = {10.1093/jamia/ocv080}
}

@article{jaeger2014,
   author = {Stefan Jaeger and Sema Candemir and Sameer Antani and Yi-Xiang J. Wáng and Pu-Xuan Lu and George Thoma},
   title = {Two public chest {X}-ray datasets for computer-aided screening of pulmonary diseases},
   journal = {Quantitative Imaging in Medicine and Surgery},
   volume = {4},
   number = {6},
   pages = {475-477},
   year = {2014},
   month = {12},
   doi = {10.3978/j.issn.2223-4292.2014.11.20}
}

@article{rsna_bone_age,
   author = {Safwan S. Halabi and Luciano M. Prevedello and Jayashree Kalpathy-Cramer and Artem B. Mamonov and Alexander Bilbily and Mark Cicero and Ian Pan and Lucas A. Pereira and Rafael T. Sousa and Nitamar Abdala and Felipe C. Kitamura and Hans H. Thodberg and Leon Chen and George Shih and Katherine Andriole and Marc D. Kohli and Bradley J. Erickson and Adam E. Flanders},
   title = {The {RSNA} Pediatric Bone Age Machine Learning Challenge},
   journal = {Radiology},
   volume = {290},
   number = {2},
   pages = {498-503},
   year = {2019},
   month = {2},
   doi = {10.1148/radiol.2018180736}
}

@article{rsna_pe,
   author = {Errol Colak and Felipe C. Kitamura and Stephen B. Hobbs and Carol C. Wu and Matthew P. Lungren and Luciano M. Prevedello and Jayashree Kalpathy-Cramer and Robyn L. Ball and George Shih and Anouk Stein and Safwan S. Halabi},
   title = {The {RSNA} pulmonary embolism {CT} dataset},
   journal = {Radiology: Artificial Intelligence},
   volume = {3},
   number = {2},
   pages = {e200254},
   year = {2021},
   month = {1},
   doi = {10.1148/ryai.2021200254}
}

@misc{rsna_cervical_2022,
   author = {Adam Flanders and Chris Carr and Errol Colak and Felipe Kitamura and Hui Ming Lin and Jeff Rudie and John Mongan and Katherine Andriole and Luciano Prevedello and Michelle Riopel and Robyn Ball and Sohier Dane},
   title = {{RSNA} 2022 Cervical Spine Fracture Detection},
   year = {2022},
   publisher = {Kaggle},
   howpublished = {\url{https://kaggle.com/competitions/rsna-2022-cervical-spine-fracture-detection}}
}

@article{rsna_abdominal_2023,
   author = {Jeffrey D. Rudie and Hui Ming Lin and Robyn L. Ball and Sabeena Jalal and Luciano M. Prevedello and Savvas Nicolaou and Brett S. Marinelli and Adam E. Flanders and Kirti Magudia and George Shih and Melissa A. Davis},
   title = {The {RSNA} Abdominal Traumatic Injury {CT} ({RATIC}) Dataset},
   journal = {Radiology: Artificial Intelligence},
   volume = {6},
   number = {6},
   pages = {e240101},
   year = {2024},
   month = {10},
   doi = {10.1148/ryai.240101}
}

@misc{rsna_lumbar_2024,
   author = {Tyler Richards and Jason Talbott and Robyn Ball and Errol Colak and Adam Flanders and Felipe Kitamura and John Mongan and Luciano Prevedello and Maryam Vazirabad},
   title = {{RSNA} 2024 Lumbar Spine Degenerative Classification},
   year = {2024},
   publisher = {Kaggle},
   howpublished = {\url{https://kaggle.com/competitions/rsna-2024-lumbar-spine-degenerative-classification}}
}

\end{document}